\documentclass[11pts]{article}
\usepackage{pdflscape}
\usepackage{amsmath,amssymb,amsthm}
\usepackage{graphicx}
\usepackage{hyperref}
\usepackage{enumitem}
\usepackage{natbib}
\usepackage{multirow}
\usepackage{array}
\usepackage{graphicx}
\usepackage{booktabs}
\usepackage{xcolor}
\usepackage{pifont}
\usepackage{comment}
\usepackage{graphicx} 
\usepackage{multirow}
\usepackage{array}
\usepackage[a4paper,
            left=2.5cm,
            right=2.5cm,
            top=3cm,
            bottom=3cm,
            includehead,
            includefoot]{geometry}

\usepackage{graphicx} 
\usepackage{multirow}
\usepackage{array}
\usepackage{pdflscape}
\usepackage{booktabs}
\usepackage{xcolor}
\newtheorem{remark}{Remark}

\newcommand{\cmark}{\textcolor{blue}{\ding{51}}} 
\newcommand{\xmark}{\textcolor{red}{\ding{55}}}   
\usepackage{tikz}
\usetikzlibrary{arrows.meta, positioning, calc}
\usepackage{caption}

\title{
Turning Time Series into Algebraic Equations: Symbolic Machine Learning for Interpretable Modeling of Chaotic Time Series
}
\author{Madhurima Panja$^{1,\dagger}$, Grace Younes$^{1,\dagger}$, and Tanujit Chakraborty$^{1,2,}$\footnote{Email: tanujit.chakraborty@sorbonne.ae} \\ $^1$SAFIR, Sorbonne University Abu Dhabi, UAE \\
$^2$Sorbonne Center for Artificial Intelligence, Sorbonne University, Paris, France\\
$^\dagger$Joint First Author}
\date{\today}
\usepackage{algorithm}
\usepackage{algpseudocode} 
\begin{document}

\maketitle

\begin{abstract}
Chaotic time series are notoriously difficult to forecast: small uncertainties in initial conditions amplify rapidly, while strong nonlinearities and regime-dependent variability constrain predictability. Although modern deep learning often delivers strong short-horizon accuracy, its black-box nature limits scientific insight and practical trust in settings where understanding the underlying dynamics matters. To address this gap, we propose two complementary symbolic forecasters that learn explicit, interpretable algebraic equations from chaotic time series data. Symbolic Neural Forecaster (SyNF) adapts a neural network–based equation learning architecture to the forecasting setting, enabling end-to-end differentiable discovery of compact and interpretable algebraic relations. The Symbolic Tree Forecaster (SyTF) builds on evolutionary symbolic regression to search directly over equation structures under a principled accuracy–complexity trade-off. We evaluate both approaches in a rolling-window nowcasting setting (one-step-ahead) using several accuracy metrics and compare against a broad suite of baselines spanning classical statistical models, tree ensembles, and state-of-the-art deep learning architectures. Numerical experiments cover a benchmark of 132 low-dimensional chaotic attractors and two real-world chaotic time series (weekly dengue incidence in San Juan and the Niño 3.4 sea-surface temperature index). Across datasets, symbolic forecasters achieve competitive one-step-ahead accuracy while providing transparent equations that reveal salient aspects of the underlying dynamics. 

\end{abstract}

\section{Introduction}
Forecasting complex dynamical systems remains one of the central challenges in modern science \cite{shukla1998predictability}. In domains ranging from epidemiology and climate to ecology and networked systems, the evolution of observable quantities is governed by nonlinear interactions, regime changes, and feedback mechanisms that are only partially understood \cite{boffetta2002predictability}. Classical forecasting approaches typically fall into two broad categories: empirical models that fit observed trajectories directly \cite{hyndman2018forecasting} and mechanistic models derived from governing differential equations \cite{brunton2022data}. While empirical models often provide short-term predictive accuracy, they may lack structural validity outside the calibration window. Conversely, mechanistic models offer interpretability and theoretical grounding but require strong prior assumptions and involve challenging inverse problems for parameter estimation \cite{tang2020introduction}.

Recent advances in data-driven modeling have sought to bridge this gap. The emerging paradigm of data-driven equation discovery aims to infer explicit symbolic representations of governing equations directly from observational data, combining the predictive flexibility of machine learning with the transparency of mathematical models \cite{song2024towards}. In contrast to black-box predictive architectures, equation discovery methods yield compact algebraic or differential expressions that are interpretable, analytically tractable, and potentially generalizable across regimes. This paradigm has gained momentum across disciplines, including physics, geosciences, and epidemiology, where uncovering hidden dynamical structure is as important as forecasting accuracy. Among modern approaches, neural-symbolic regression has emerged as a powerful strategy for discovering governing equations in high-dimensional and networked systems \cite{hu2025learning}. By combining neural networks for nonlinear representation learning with symbolic regression for equation parsing, these methods reduce dimensionality while retaining interpretability. However, despite rapid progress in discovering governing equations of dynamical systems, the role of symbolic regression in time-series forecasting, particularly under chaotic dynamics, remains underexplored. Most existing works focus on recovering differential equations or static functional relationships, rather than evaluating predictive performance in rolling forecasting settings.

Chaotic systems present a particularly stringent challenge for forecasting methodologies \cite{kantz2003nonlinear}. Their sensitive dependence on initial conditions limits long-horizon predictability, while nonlinear interactions and regime-dependent behaviors restrict the use of conventional linear or purely data-driven models \cite{boffetta2002predictability}. In practice, forecasters often observe only partial system states, further complicating equation recovery. This raises a critical question: {\it Can symbolic regression discover compact, interpretable forecasting equations that remain competitive with modern machine learning models in chaotic time-series prediction?} Furthermore, chaotic forecasting is not merely a theoretical challenge but a problem of substantial practical importance. Applications range from predicting the El Niño–Southern Oscillation \cite{allan1996nino} and tracking dengue fever outbreaks \cite{benedum2020weekly} to anticipating rogue ocean waves \cite{bonatto2011deterministic} and monitoring instability in financial markets \cite{kuzmenko2023dynamic}. In each of these domains, uncertainty amplifies rapidly, and nonlinear feedback mechanisms dominate system evolution. Improving short-term predictive accuracy in such settings, therefore, carries significant societal and economic implications, particularly when forecasts inform public health interventions and climate risk management.

Modern data-driven approaches such as reservoir computing \cite{jaeger2004harnessing} and deep recurrent networks \cite{hochreiter1997long} have demonstrated strong short-term forecasting ability in chaotic settings, yet they remain largely ``black-box'' models. While effective locally, their long-horizon forecasts deteriorate rapidly and their performance is often sensitive to noise, data scarcity, and hyperparameter tuning. Even recent long-sequence architectures, including Transformers and their time-series variants \cite{vaswani2017attention, zhou2021informer}, demand substantial computational resources while offering zero structural interpretability. To address these concerns, methods grounded in dynamical systems theory have been explored. Phase-space reconstruction, based on Takens’ embedding theorem and its extensions \cite{takens2006detecting, sauer1991embedology,deyle2011generalized}, reconstructs attractor geometry through delay embeddings, but may struggle with complex nonlinear couplings, even with refinements such as Ma's inverse delayed embedding \cite{strogatz2018nonlinear}. Analytical indicators can provide early-warning signals when governing equations are known \cite{dakos2008slowing,boffetta2002predictability}, yet lack general applicability, while vector-field reconstruction estimates local flow directly from data but deteriorates under sparse or noisy observations \cite{farmer1987predicting}. These strands reveal a persistent tension: highly accurate models often lack interpretability, whereas transparent dynamical techniques can fail in complex, data-limited environments. 

Recent advances in sparse system identification, notably SINDy \cite{brunton2016discovering}, and perturbation-based learning approaches \cite{liu2024interpretable,enguehard2023learning}, demonstrate that compact, closed-form equations can be inferred directly from data, rekindling interest in symbolic techniques for interpretable forecasting. Symbolic regression (SR) provides a natural resolution to this problem by jointly minimizing prediction error and model complexity, thereby uncovering concise formulas that both fit observed data and expose the underlying functional relationships \cite{koza1994genetic, schmidt2009distilling}. Although the search space is NP-hard \cite{song2024prove, virgolin2022symbolic} and computationally intensive, SR offers several important advantages: it does not impose a predefined model structure, enabling the discovery of novel functional forms; it yields compact, algebraic equations that enhance interpretability \cite{kotanchek2013symbolic, vladislavleva2008order}; and it often exhibits strong generalization, particularly in data-scarce regimes \cite{wilstrup2021symbolic}. Despite recent methodological advances, symbolic regression has not been systematically benchmarked for chaotic time-series forecasting across a broad and diverse set of systems. In particular, its performance relative to modern forecasting architectures remains largely unexplored. 

We address this gap by assembling a curated benchmark of 132 low-dimensional chaotic attractors, including classical systems such as Lorenz and Rössler, following \cite{gilpin2024generative, gilpin2021chaos, gilpin2023model}. In this work, we adapt symbolic frameworks to discrete time-series forecasting, enabling the extraction of interpretable dynamical relations directly from lagged observations. We propose two complementary symbolic forecasters that learn explicit algebraic mappings from past observations to future states. The first is a neural-symbolic architecture (namely, \emph{Symbolic Neural Forecaster (SyNF)}) that extends the Equation Learner (EQL) framework \cite{sahoo2018learning} to the forecasting setting through end-to-end differentiable training. The second approach namely \emph{Symbolic Tree Forecaster (SyTF)} builds on the PySR symbolic regression library \cite{cranmer2023interpretable} and employs evolutionary symbolic regression to search over expression trees under an explicit accuracy–complexity trade-off. By reframing symbolic regression as a forecasting methodology rather than solely a scientific discovery tool, this work advances the development of forecasting models that are not only competitive in accuracy but also transparent and mechanistically meaningful. We evaluate both methods on a nowcasting task (one-step-ahead prediction) and compare their performance against a broad suite of baselines, including deep-learning architectures and tree-based regressors using several error metrics. To evaluate real-world applicability, we additionally include two empirical time series: the El Niño SST index \cite{ray2021optimized} and weekly dengue cases \cite{chakraborty2019forecasting} in San Juan.

The rest of this paper is structured as follows. Section 2 reviews background material on symbolic regression. Section 3 introduces our benchmarking datasets: 132 low-dimensional chaotic attractors (e.g., Lorenz, Rössler, Chua) along with two real-world series (the El Niño 3.4 index and dengue cases in San Juan). Section 4 presents our proposed methodology. Section 5 outlines our experimental evaluation on both synthetic and real-world datasets. Finally, Section 6 concludes the paper with a discussion on the limitations and future directions of this study.






\section{Background: Symbolic Regression}

Symbolic Regression (SR) is an emerging subfield of machine learning that aims to discover symbolic mathematical expressions describing the relationships within data, wherein both the parameters and the functional structure of the analytical model are simultaneously optimized \cite{schmidt2009distilling, makke2024interpretable}. Unlike conventional deep learning architectures, SR emphasizes on the trade-off between predictive accuracy and interpretability. A model is said to be interpretable if the relationship between model inputs and outputs can be traced concisely through structured analytical equations \cite{la2021contemporary}. As data-driven methodologies are increasingly applied in scientific disciplines and high-stakes decision-making, the need for interpretable models has become more critical. In domains such as physics, interpretable mathematical models grounded in first principles facilitate reasoning about the underlying mechanisms in ways that opaque predictive models, such as deep neural networks, cannot. Similarly, in high-stakes decision fields such as public health, the deployment of accurate, non-interpretable models remains limited due to safety and accountability concerns \cite{virgolin2020machine}.


Consider a dataset $\mathcal{D}=\left\{\left(\mathbf{x}_i, z_i\right)\right\}_{i=1}^N$, where $\mathbf{x} \in \mathbb{R}^{q}$ denotes the input features and $z$ represents the target. A linear model $z=\theta^{\top} \mathbf{x}$ is fully interpretable, since the contribution of each input can be captured from its coefficient. However, this model is often structurally restrictive in capturing complex nonlinear patterns. In contrast, a neural network $z=\mathrm{NN}(\mathbf{x}; \theta)$ can model complex relationships but remains opaque as the mapping from $\mathbf{x}$ to $z$ remains entangled within multiple layers of learned transformations. SR addresses this gap by seeking a mapping $\hat{z}(\mathbf{x})=\hat{\alpha}(\mathbf{x}, \hat{\beta})$ that is both accurate and interpretable. By assuming an underlying analytical relation $z(\mathbf{x})=\alpha^*\left(\mathbf{x}, \beta^*\right)+\epsilon, \mathrm{SR}$ jointly explores the space of candidate expressions $\alpha^*$ and their associated parameters $\beta^*$. In this way, SR directly targets the recovery of an explicit functional form that reflects the true structure of the data-generating process, achieving interpretability through model transparency rather than post-hoc explanation \cite{sabbatini2025four}.

A variety of algorithmic frameworks have been developed to make the SR search process computationally feasible and scalable. These methods can be broadly classified into four categories: exhaustive (brute-force) search, genetic programming, sparse regression, and deep learning–based frameworks, each with distinct strengths and limitations. The exhaustive search approaches, such as BACON \cite{langley1977bacon} and FAHRENHEIT \cite{langley1989data}, systematically generate and evaluate candidate expressions, successfully rediscovering empirical laws from idealized datasets. However, their reliance on heuristics and the exponential growth of the search space make brute-force methods infeasible for high-dimensional, complex nonlinear problems. The Genetic programming (GP)–based SR methods iteratively evolve candidate expressions using mutation, crossover, and selection \cite{koza1994genetic}. Modern variants, including PySR \cite{cranmer2023interpretable}, GP-GOMEA \cite{virgolin2021improving}, and QLattice \cite{brolos2021approach}, can explore large, unstructured model spaces and uncover complex nonlinear relationships. Despite their flexibility, GP-based methods often suffer from expression bloat, are computationally intensive in high dimensions, and are sensitive to hyperparameters, which limits scalability in real-world applications. On the other hand, the sparse regression approaches assume that systems can be described using a few active terms. Exact methods using Mixed-Integer Nonlinear Programming (MINLP) \cite{cozad2018global} guarantee sparsity but quickly become intractable as the feature library (in terms of size of the feature space and number of mathematical operations) grows. More scalable $\ell_1$-regularized variants (e.g., LASSO, elastic net) promote sparsity efficiently but often retain unnecessary terms. The Sparse Identification of Nonlinear Dynamics (SINDy) framework \cite{brunton2016discovering} extends sparse regression to dynamical systems by fitting coefficients in a predefined function library; while fast and effective for structured dynamics, its performance is sensitive to noise and the selection of the library. The Sure Independence Screening and Sparsifying Operator (SISSO) \cite{ouyang2018sisso} and SyMANTIC \cite{muthyala2025symantic} methods improve scalability and interpretability through feature screening, recursive expansion, Pareto-front tracking, and GPU acceleration \cite{muthyala2024torchsisso}, but they remain limited to expressions within the predefined feature set. The deep learning–inspired SR approaches embed symbolic discovery within neural architectures. Recently developed models such as Deep Symbolic Optimization \cite{petersen2019deep}, Equation Learner (EQL) \cite{sahoo2018learning}, and SR-Transformer \cite{kamienny2022end} optimize both structure and parameters using differentiable networks and sequence-to-sequence models. Hybrid methods like AI-Feynman \cite{udrescu2020ai} combine neural fitting with targeted searches informed by domain knowledge. While powerful for capturing complex patterns, these approaches often require large datasets for effective training and can struggle with low-data regime problems. Thus, symbolic regression has emerged as a powerful tool for uncovering governing equations across diverse scientific domains, including physics, biochemistry, ecology, epidemiology, and complex network dynamics \cite{hu2025learning}. By automatically extracting interpretable mathematical relations from rich observational data, it reveals hidden mechanisms underlying phenomena such as epidemic transmission, social interaction patterns, and nonlinear system evolution, thereby supporting more transparent modeling and informed decision-making \cite{fokas2023algebraic}.

The choice of modeling approach is often guided by the structure of the problem and the degree of prior knowledge available \cite{song2024towards}. When the objective is to discover complex, nonlinear equations with minimal prior assumptions, symbolic regression provides a flexible framework for uncovering explicit mathematical relationships directly from data \cite{hu2025learning}. In contrast, when substantial structural knowledge is available, such as when seeking to identify governing ODEs or PDEs, sparse regression methods offer a principled way to recover parsimonious dynamical systems. Finally, for problems involving high-dimensional interactions and rich nonlinear dependencies, equation learner networks provide a scalable neural-symbolic alternative that balances expressive power with structural interpretability. These variations in SR methods provide the context for symbolic forecasting approaches that leverage the strengths of GP-based algorithms, such as PySR, and neural-symbolic architectures like EQL, to discover interpretable and accurate models for sequential learning tasks in time series datasets.

\section{Motivating Examples}\label{sec:datasets}

In this study, we evaluate the performance of symbolic forecasting approaches relative to baseline architectures by assessing their modeling and forecasting capabilities across a diverse set of datasets. We consider two distinct categories of data: synthetic time series generated from chaotic dynamical systems and complex real-world observations drawn from applied domains. This dual evaluation strategy allows us to demonstrate both the theoretical robustness of the symbolic forecasters and their practical relevance. In this section, we briefly describe these two distinct data categories and analyze their global characteristics.


\subsection{Synthetic Dataset from Chaotic Attractors}\label{Synthetic_data_Description}

\begin{figure}[htbp]
  \centering
  \includegraphics[width=\linewidth]{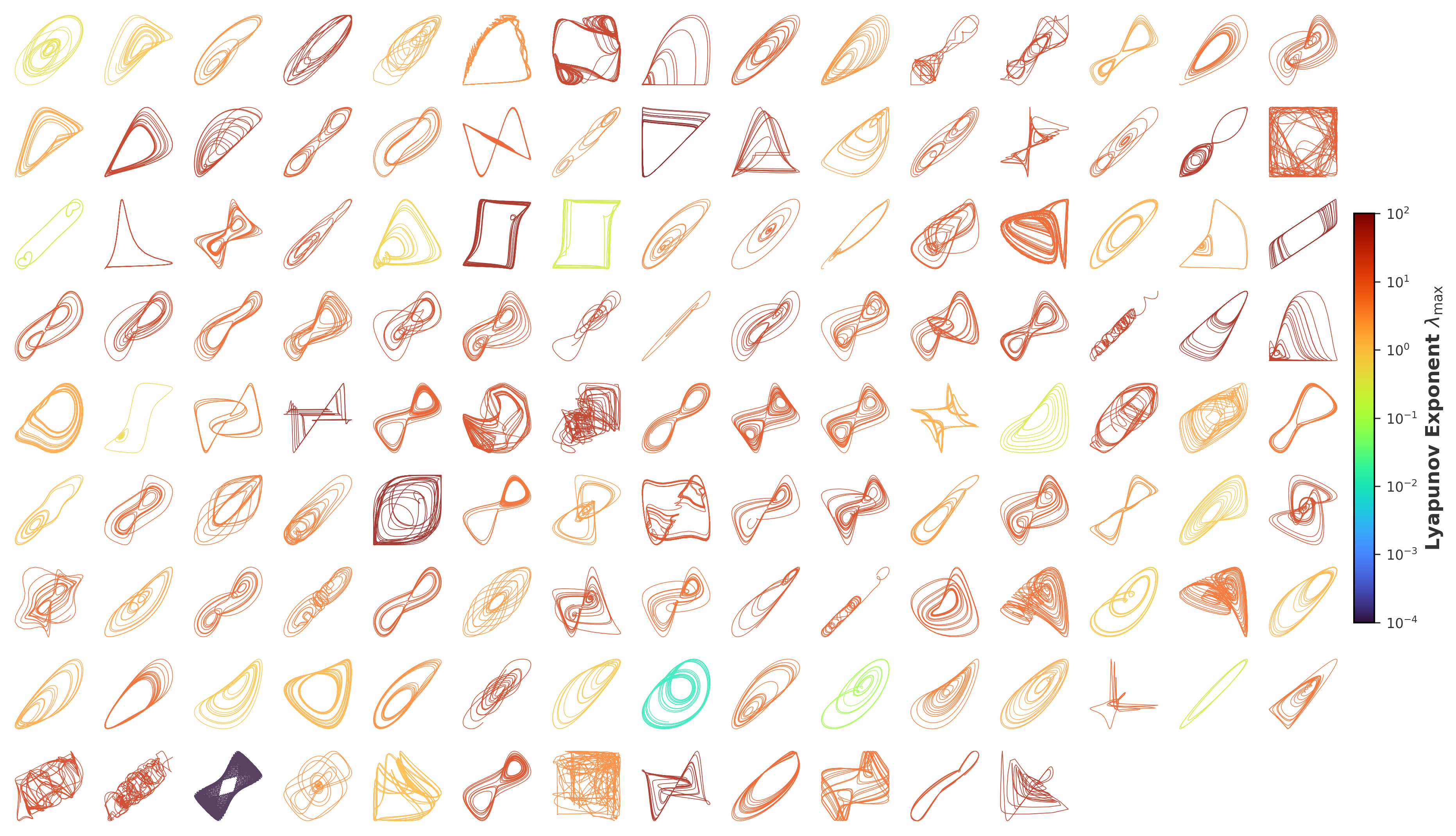}
  \caption{A dataset of 132 distinct low-dimensional chaotic systems, colored by largest Lyapunov exponent ($\lambda_{\max}$).}
  \label{fig:attractor_grid}
\end{figure}

In our analysis, we consider 132 low-dimensional chaotic system datasets collected from the publicly available `dysts' repository \cite{gilpin2021chaos, gilpin2023model}. These dynamical systems encompass a wide range of applied domains, including climatology, neuroscience, astrophysics, ecology, and others, thereby providing a diverse benchmark for evaluating forecasting performance on simulated observations. Each dataset is univariate and corresponds to a fine-grained temporal granularity, with 100 points sampled per period. The time series for each system is obtained by numerically integrating the canonical differential equations that define the underlying chaotic dynamics. The parameters and initial conditions are selected based on the literature to ensure sustained chaotic behavior \cite{gilpin2021chaos}. By slightly perturbing the initial conditions for each system, we capture the intrinsic sensitivity to initial states characteristic of chaotic systems, resulting in diverse yet structurally consistent trajectories. These attractors provide a controlled yet complex benchmark for evaluating forecasting and data-driven modeling approaches. In our study, we simulate chaotic attractor series with 1200 observations, and we split each series chronologically into the first 1000 observations for training and the remaining 200 observations for testing. This preserves temporal dependencies and avoids data leakage during forecast evaluation.

Furthermore, to characterize the chaotic behavior of each simulated data we compute the largest Lyapunov exponent ($\lambda_{\max}$). This exponent quantifies the average exponential rate of divergence between nearby trajectories in the system’s state space. A positive value of $\lambda_{\max}$ indicates chaotic dynamics, while a value of zero corresponds to neutral stability, often associated with quasi-periodicity, and a negative value represents stable behavior where trajectories converge to fixed points. Figure \ref{fig:attractor_grid} provides a visual representation of all the dynamical systems, color-coded according to the values of the largest Lyapunov exponent ($\lambda_{\max}$). As shown in the figure, all synthetic datasets have positive values of $\lambda_{\max}$, indicating that they exhibit significant chaotic behavior.

\subsection{Real‐World Time Series}\label{Sec_Real_World_Time_Series}
Alongside synthetic datasets from chaotic attractors, we consider two real-world datasets from the epidemic and climate domains, namely the San Juan dengue cases and El Niño sea‐surface temperature (SST). The San Juan dengue dataset comprises the total number of dengue infections reported weekly in the Puerto Rico region from 1990 to 2013. This dataset has been widely used in epidemic forecasting studies due to its complex structural patterns and strong seasonal cycles \cite{johansson2019open, panja2023epicasting}. The El Niño SST dataset characterizes the El Niño–Southern Oscillation, a recurrent disruption in the equatorial Pacific Ocean driven by interactions between atmospheric and oceanic circulations and marked by elevated sea-surface temperatures \cite{ray2021optimized}. This phenomenon profoundly influences global climate variability and marine ecosystems, making accurate SST forecasting a major challenge in climate science. In our analysis, we consider weekly SST data recorded in the Niño 3.4 region spanning January 3, 1990, to April 21, 2021. 

\begin{figure}
    \centering
    \includegraphics[width=1.0\linewidth]{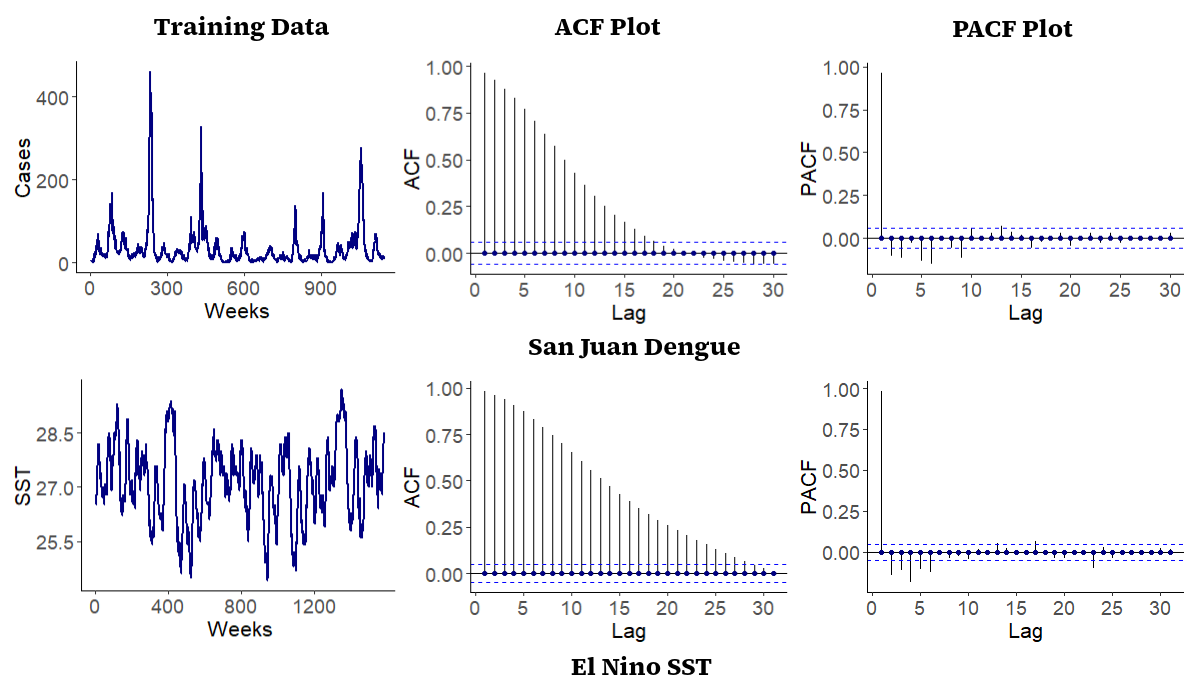}
    \caption{Real-world datasets along with their autocorrelation function (ACF) and partial autocorrelation function (PACF).}
    \label{table_acf_pacf}
\end{figure}

In our study, we perform several preliminary analyses to examine the global behavior of these real-world datasets. To evaluate the performance of the symbolic forecasting approaches, each dataset is split chronologically into training and testing subsets. For the San Juan dengue dataset, the first 1,144 observations are used for training and the remaining 52 observations for testing, while for the El Niño SST dataset, 1,582 observations are allocated for training and 52 for testing. 
Figure \ref{table_acf_pacf} visualizes the temporal patterns of these real-world datasets along with their autocorrelation function (ACF) and partial autocorrelation function (PACF). As highlighted in the plot, the ACF and PACF functions reveal distinct underlying dynamics for these datasets. For the San Juan dengue data, the ACF decays rapidly to near zero after a few lags, while the PACF displays a significant correlation at lag 1, followed by negligible values. This pattern indicates that the epidemic incidence series exhibits short-memory behavior, with only limited dependence on past observations. In contrast, the El Niño SST dataset shows a slowly decaying, sinusoidal ACF pattern consistent with seasonal or cyclic dynamics, while its PACF again highlights a strong influence at lag 1 and weaker effects at higher lags. 

To analyze the global characteristics of these real-world datasets, we examine several statistical features, namely non-stationarity, nonlinearity, seasonality, normality, and chaotic dynamics. These features offer comprehensive insights into the underlying structure and dynamical patterns of the time series. To evaluate nonlinearity, we employ Terasvirta’s neural network test, which examines the null hypothesis that the observed time series follows a linear structure. Stationarity is another fundamental property of a time series, which guarantees that the statistical features, such as mean and variance, of the series do not change over time. In our analysis, to assess the stationarity of a time series, we employ the Kwiatkowski–Phillips–Schmidt–Shin (KPSS) test. We also examine seasonality, which captures recurring patterns or periodic fluctuations at fixed intervals, using Ollech and Webel’s combined test. Additionally, normality of the time series is assessed with the Anderson-Darling test, which determines whether the series follows a Gaussian distribution, with deviations indicating skewness, heavy tails, or other structural irregularities. The results of these statistical tests, reported in Table \ref{Table_Data_Char_real}, suggest that both the observed series exhibit stationary patterns, chaotic behavior, and significantly deviate from normality assumptions. The San Juan dengue dataset exhibits nonlinear dynamics with seasonal fluctuations at regular intervals, while the El Niño SST dataset provides linear trends coupled with seasonal variations. 
Overall, these observations highlight the inherent temporal structures across both real-world datasets and underscore the need for developing forecasting methods that can effectively balance accuracy with interpretability.

\begin{table}[h]
\centering
\caption{Statistical characteristics of real-world datasets. In the table, a \cmark$\;$denotes the presence of a feature, while a \xmark$\;$denotes its absence. Values in parentheses represent the p-values from the corresponding statistical tests and the maximum Lyapunov exponent.}
\scriptsize
\begin{tabular}{lcccccccc}
\toprule
Data & Time Span & Length & Granularity & Linearity & Stationarity & Normal & Seasonality & Chaotic \\
\midrule
Sanjuan Dengue & 1990--2013 & 1196 & Weekly & \xmark$\;$(0.026) & \cmark$\;$(0.098) & \xmark$\;$(0.001) & \cmark$\;$(0.001) & \cmark$\;$(0.001) \\
El Niño SST & 1990--2021 & 1634 & Weekly & \cmark$\;$(0.777) & \cmark$\;$(0.100) & \xmark$\;$(0.001) & \cmark$\;$(0.001) & \cmark$\;$(0.001) \\
\hline
\end{tabular}
\label{Table_Data_Char_real}
\end{table}

\section{Symbolic Forecasting Approaches}


This section provides the formulation of the symbolic forecasting techniques, which aim to combine strong predictive performance with model interpretability for time series forecasting. We introduce two distinct approaches, namely the SyNF and SyTF, based on the neural training and evolutionary search techniques of the symbolic regression models. SyNF framework adopts the EQL architecture \cite{martius2016extrapolation} to time series forecasting, where the conventional neural network activations (e.g., ReLU, tanh) are replaced with a set of unary and binary mathematical operations. These operations serve as symbolic building blocks that the model combines into larger expressions as training progresses. Since the entire architecture remains differentiable, the network can be trained end-to-end while simultaneously producing explicit, interpretable equations expressed in terms of lagged input features. 

\begin{figure}
    \centering
    \includegraphics[width=1.0\linewidth]{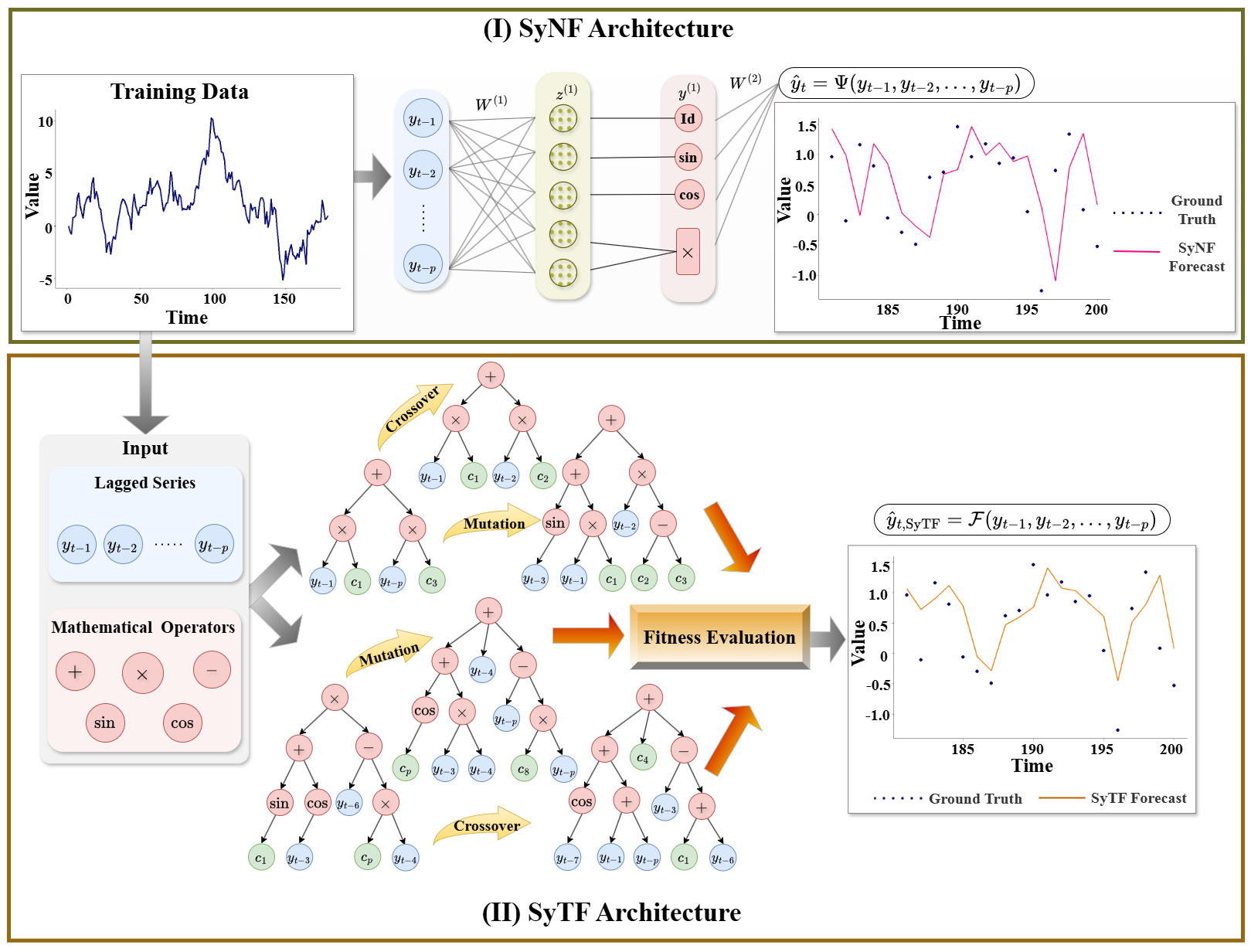}
    \caption{\textbf{Symbolic Forecasting Architectures.} The upper panel illustrates the architecture of the Symbolic Neural Forecaster (SyNF), where a differentiable neural training mechanism is used to learn interpretable forecasting equations from symbolic operators and lagged inputs. The lower panel presents the workflow of the Symbolic Tree Forecaster (SyTF), which employs an evolutionary search to construct and refine symbolic expression trees based on predictive performance and interpretability. The demonstration is conducted on a simulated dataset. The mathematical expression obtained from the SyNF model is $\hat{y}_t = 0.1110 + 0.9421y_{t-1} + 0.0001\left(y_{t-1}\right)^2$ (top image) whereas the SyTF architecture yields $\hat{y}_{t, \text{SyTF}} = 0.9613y_{t-1}$ (bottom image) using a single historical lagged observation. The one-step-ahead rolling window forecasts are also presented in the upper and lower panels for the simulated example datasets.}    
    \label{Fig_SyNF_SyTF}
\end{figure}

Our second approach, the genetic-programming-based SyTF, builds on the PySR framework \cite{cranmer2023interpretable}. SyTF employs an evolutionary search process inspired by natural selection, where candidate mathematical expressions are constructed from a predefined library of basic operations, such as addition, subtraction, multiplication, division, and elementary nonlinear or trigonometric functions. These expression trees are iteratively mutated, recombined, and selected based on their predictive accuracy and model complexity. Similar to SyNF, SyTF models temporal dependencies through lagged inputs, but unlike the differentiable architecture of SyNF, it explores a broad, unstructured model space using evolutionary operators without relying on gradient-based training. Together, SyNF and SyTF offer two distinct yet complementary perspectives on symbolic forecasting. Figure \ref{Fig_SyNF_SyTF} illustrates the schematic architecture of the symbolic forecasting approaches. The upper panel (I) depicts the SyNF framework, where a neural optimization-driven mechanism is utilized for discovering interpretable predictive expressions, while the lower panel (II) demonstrates the workflow of the SyTF model and reflects the strengths of evolutionary exploration in navigating large symbolic search spaces. Through these two distinct approaches, we demonstrate how symbolic forecasting frameworks can yield accurate, transparent, and interpretable models across both synthetic chaotic systems and real-world datasets. 

\subsection{Symbolic Neural Forecaster}

Given an univarite input series $\left\{y_{1}, y_{2},\ldots, y_{T}\right\}$, the SyNF framework assumes that the observation at time $t$ can be expressed as 
$$
y_{t} = \Phi\left(\underline{y}_{t-1}\right) + \xi_{t}, 
$$
where $\underline{y}_{t-1} = \left\{y_{t-1}, y_{t-2}, \ldots, y_{t-p}\right\}^{\top} \in \mathbb{R}^p$ contains the $p$-lagged observations, $\Phi : \mathbb{R}^p \rightarrow \mathbb{R}$ denotes the true underlying functional mapping, and $\xi_{t}$ is the additive noise. The objective is to learn a compact and interpretable approximation $\Psi: \mathbb{R}^p \rightarrow \mathbb{R}$ that minimizes the expected squared error $\mathbb{E}\left[\left(\Psi\left(\underline{y}_{t-1}\right) - \Phi\left(\underline{y}_{t-1}\right)\right)^2\right]$. Since the expected loss is unknown, we estimate it using the empirical risk function $\frac{1}{T-p}\sum_{t = p+1}^T \left(\Psi\left(\underline{y}_{t-1}\right) - y_t\right)$. The resulting analytical approximation $\Psi(\cdot)$ provides one-step ahead forecast $(\hat{y}_{_{T+1}})$ of the observed series 
$$
\hat{y}_{_{T+1}} = \Psi\left(\underline{y}_{T}\right).
$$
To learn $\Psi(\cdot)$ from historical data, SyNF modifies a standard feed-forward neural network by replacing conventional activations with fixed algebraic base functions, including sine, cosine, identity, and multiplication \cite{sahoo2018learning}. These operations act as symbolic building blocks that can be combined within a differentiable architecture, enabling gradient-based training while still producing interpretable expressions. SyNF employs a single hidden-layer architecture. A linear transformation is applied for mapping the lag-embedded input $\underline{y}_{t-1} \in \mathbb{R}^p$ to a $d$-dimensional intermediate representation 
$$
z_{t-1}^{(1)} = W^{(1)} \underline{y}_{t-1} + b^{(1)},
$$
where $W^{(1)} \in \mathbb{R}^{d \times p}$ and $b^{(1)} \in \mathbb{R}^{d}$ are learnable parameters. The hidden layer consists of $u$ unary functions $f_i : \mathbb{R} \rightarrow \mathbb{R}$ and $v$ binary functions $g_j : \mathbb{R}\times\mathbb{R} \rightarrow \mathbb{R}$, applied to the components of $z^{(1)}_{t-1} = [z_{t-1,1}, \ldots, z_{t-1,d}]$ where $d = u + 2v$. The unary operations process the first $u$ components, while binary operations process the remaining components in pairs. 
The hidden output ($y_{t-1}^{(1)} \in \mathbb{R}^k$) is obtained by concatenating the symbolic outputs as 
\begin{equation}\label{Eq_synf}
    y_{t-1}^{(1)} := \left[f_1(z_{t-1, 1}), f_2(z_{t-1, 2}), \ldots, f_u(z_{t-1, u}),  g_1(z_{t-1, u+1}, z_{t-1, u+2}), \ldots, g_v(z_{t-1, d-1}, z_{t-1, d})\right]^{\top},
\end{equation}
where $k = u + v$. In the SyNF framework to enhance model interpretability and robustness, we utilize three unary functions, i.e., $f_i \in \left\{\operatorname{identity}, \;\operatorname{sine}, \text{ and } \operatorname{cosine}\right\}$ and the binary operations represent the element-wise multiplication. The one-step ahead prediction is generated from the output layer by applying a linear transformation on the hidden activation $y_{t-1}^{(1)}$ as:
$$
\hat{y}_t = W^{(2)} y_{t-1}^{(1)} + b^{(2)},
$$
where $W^{(2)} \in \mathbb{R}^{1 \times k}$, $b^{(2)} \in \mathbb{R}$ are learnable parameters of the output layer. Hence, by stacking several symbolic units, SyNF effectively constructs weighted combinations of elementary algebraic terms, enabling interpretable equation discovery while maintaining strong predictive capability. The base model of the SyNF architecture is trained by minimizing the mean squared error (MSE)
$$\operatorname{MSE} = \frac{1}{T-p}\sum_{t=p+1}^T(y_t - \hat{y}_t)^2$$ 
which is differentiable in its parameters $\left\{W^{(1)}, b^{(1)}, W^{(2)}, \text{ and } b^{(2)}\right\}$. The training is performed using a gradient descent backpropagation approach with the Adam optimizer, without regularization. To prevent overfitting and promote parsimony, we additionally introduce a regularized variant, SyNF-Reg, which augments the MSE loss function with an $\ell_1$ penalty
$$\operatorname{MSE} + \lambda \left(\left\|W^{(1)}\right\|_1 + \left\|W^{(2)}\right\|_1\right),$$
where $\lambda$ controls the sparsity level and encourages the removal of redundant symbolic components.

\paragraph{SyNF-Div Framework.} While the SyNF architecture captures a broad class of nonlinear relationships through additive, multiplicative, and trigonometric operations, many real-world and dynamical systems inherently rely on division-based functional forms, such as rational dependencies, saturation dynamics, and feedback mechanisms. To model such behavior, we extend the SyNF framework by introducing learnable division units, following the design principles of the EQL$^{\div}$ model \cite{sahoo2018learning}. A critical challenge in introducing division operations is the numerical instability that occurs when denominators approach zero. The SyNF-Div family addresses this by penalizing small denominators and gradually relaxing this constraint during training, allowing the network to learn meaningful rational structures while maintaining stable optimization. This results in two variants: SyNF-Div, which incorporates division into the symbolic function set, and SyNF-Div-Reg, which further integrates $\ell_1$ regularization to promote sparsity and interpretability.

The SyNF-Div framework retains the one-hidden-layered feed-forward architecture of the SyNF model, with the output layer augmented by a division operation. Specifically, the hidden activation $y_{t-1}^{(1)}$ (as in Eq. \ref{Eq_synf}) is linearly transformed as:
$$
z_{t-1}^{(2)} = W_*^{(2)} y_{t-1}^{(1)} + b_*^{(2)},
$$
where $z_{t-1}^{(2)} = \left[z_{t-1, 1}^{(2)}, z_{t-1, 2}^{(2)}\right] \in \mathbb{R}^2$ and $\left\{W_*^{(2)} \in \mathbb{R}^{2\times k}, b_*^{(2)} \in \mathbb{R}^{2}\right\}$ are the learnable parameters. The one-step ahead prediction is generated from the SyNF-Div model by applying the division operation ($h_{\gamma}$) on $z_{t-1}^{(2)}$ as:
\begin{equation}\label{Eq_SYNF_DIV}
\hat{y}_{t, {*}} = h_{\gamma} \left(z_{t-1, 1}^{(2)}, z_{t-1, 2}^{(2)}\right) = \begin{cases}
\dfrac{z_{t-1, 1}^{(2)}}{z_{t-1, 2}^{(2)}}, & z_{t-1, 2}^{(2)} > \gamma \\[6pt]
0, & z_{t-1, 2}^{(2)} \leq \gamma,
\end{cases}
\end{equation}
where $\hat{y}_{t, {*}}$ is the forecast generated by the SyNF-Div framework and $\gamma \geq 0$ is a threshold controlling numerical stability, such that if the denominator falls below $\gamma$, the output and its gradient are set to zero. Additionally, to prevent division by zero, a penalty term is added that penalizes small or negative denominators. For each denominator activation $z_{t-1, 2}^{(2)}$, a local penalty is defined as:
\begin{equation}
p^\gamma\bigl(z_{t-1, 2}^{(2)}\bigr) := \max(\gamma - z_{t-1, 2}^{(2)},\,0),
\label{eq:divpenalty}
\end{equation}
where $\gamma$ is the stability threshold as in Eq.~\eqref{Eq_SYNF_DIV}. This penalty term is activated when the denominator falls below $\gamma$, and summing it over all the training samples yields the global penalty as
\begin{equation}
P^\gamma = 
\sum_{t = p+1}^T
p^\gamma\!\bigl(z_{t-1, 2}^{(2)}\bigr). 
\label{eq:globalpenalty_1}
\end{equation}
While Eq.~\eqref{eq:globalpenalty_1} prevents negative denominators during training, additional constraints are required to ensure stable behavior under out-of-sample and extrapolative regimes. To control instability arising from negative denominators or high magnitudes of output, \emph{penalty epochs} are introduced at fixed intervals (every 50 epochs), during which the model is trained using an additional penalty loss ($\mathcal{L}^{\text{Penalty}}$) given by:
\begin{equation}
\mathcal{L}^{\text{Penalty}} = P^\gamma + P^{\text{bound}}.
\label{eq:penaltyloss}
\end{equation}
The $P^{\text{bound}}$ enforces bounded outputs and is given by:
\begin{equation}
P^{\text{bound}} = \sum_{t = p+1}^T \left[\max\left(\Psi_*\left(\underline{y}_{t-1}\right) - B,\, 0\right) + \max\bigl(-\Psi_*\left(\underline{y}_{t-1}\right) - B,\, 0\bigr) \right],
\label{eq:boundpenalty_1}
\end{equation}
where $\Psi_*\left(\underline{y}_{t-1}\right) = \hat{y}_{t, {*}}$ denotes the network’s predicted output at time $t$. Here, $B$ is a user-defined bound on the output magnitude, chosen based on the range of training data, with results largely insensitive to its exact value. Furthermore, to balance numerical stability and the accurate approximation of symbolic expressions, the threshold $\gamma$ is gradually relaxed during training, with $\gamma(t)=1/\sqrt{t+1}$. This curriculum ensures safe optimization in early stages while progressively approximating exact division. During validation and testing, $\gamma$ is fixed to a small value ($10^{-4}$).

The SyNF-Div framework is fully differentiable in its parameters $\{W^{(1)}, W_*^{(2)}, b^{(1)}, b_*^{(2)}\}$ which allows network training through gradient descent backpropagation approach. The training mechanism for SyNF-Div combines predictive accuracy with global penalty term:
$$
\mathcal{L}_{\text{SyNF-Div}} = \frac{1}{T-p}\sum_{t=p+1}^T(y_t - \hat{y}_{t, {*}})^2 + P^\gamma,
$$
while SyNF-Div-Reg framework further integrates an $\ell_1$ regularization to promote sparsity:
$$
\mathcal{L}_{\text{SyNF-Div}} + \lambda_*\bigl(\|W^{(1)}\|_1 + \|W_*^{(2)}\|_1\bigr).
$$
The bound penalty term $P^{\text{bound}}$ (Eq.~\eqref{eq:boundpenalty_1}) is used in the training mechanism of both SyNF-Div and its regularized variant only during specific penalty epochs. Overall, the SyNF-Div architecture enables learning of accurate, stable, and interpretable rational structures, extending the symbolic forecasting capabilities of SyNF to a wider range of real-world systems.

\subsection{Symbolic Tree Forecaster}
The SyTF framework is a multi-population evolutionary algorithm that adopts an evolve-simplify-optimize loop inspired by the PySR architecture \cite{cranmer2023interpretable}, specifically for time series forecasting. Given the lagged inputs $\underline{y}_{t-1} = \left\{y_{t-1}, y_{t-2}, \ldots, y_{t-p}\right\}^{\top} \in \mathbb{R}^p$, the framework build an interpretable, expression tree-based symbolic function ($\mathcal{F}$) to generate one-step ahead forecast of the observed time series as follows:
$$
\hat{y}_{t, \text{ SyTF}} = \mathcal{F}\left(\underline{y}_{t-1}\right).
$$
The symbolic mapping $\mathcal{F}: \mathbb{R}^p \rightarrow \mathbb{R}$ is obtained through evolutionary search over a structured space of mathematical expressions. In SyTF, the candidate forecasting functions are represented as expression trees composed of lagged variables, scalar constants, and a predefined set of unary and binary operators (e.g., addition, subtraction, multiplication, division, $\sin$, $\cos$, or $\exp$). These trees evolve across multiple asynchronously interacting subpopulations, which helps in preserving diversity during the search. 
Within each subpopulation, tournament selection \cite{brindle1980genetic, goldberg1991comparative} is used to choose parents, favoring fitter expressions while maintaining a non-zero probability of selecting weaker candidates. The selected parent is cloned and modified through mutation and crossover operations, and the resulting offspring replaces the oldest candidate in a subpopulation. This age-regularized replacement prevents the population from stagnating and reduces the risk of early convergence to local optima \cite{real2020automl}. 

To further enhance exploration, SyTF integrates a simulated annealing mechanism during mutation \cite{cranmer2023interpretable}. Under this scheme, a structurally modified candidate with fitness $L_F$ may replace its parent with fitness $L_E$ according to the rejection probability

$$
\mathcal{P} = \exp\!\left(\frac{L_F - L_E}{\alpha\,\tau}\right),
$$
where $\tau \in [0,1]$ is a temperature parameter and $\alpha$ controls the scale of temperature. This strategy allows the evolution to alternate between high temperature and low temperature phases, with the high $\tau$ increasing diversity of individuals and low $\tau$ narrowing in on the fittest individuals. Alongside the simulated annealing strategy, SyTF performs an evolve–simplify–optimize cycle on the evolutionary process. In the evolve phase, new candidate equations are generated through repeated application of tournament selection and mutation. The simplify phase then applies algebraic identities to reduce expressions into more compact yet mathematically equivalent expressions, constraining the search without discarding useful intermediate structures. 
In the optimize phase, real-valued constants within the symbolic expressions are fine-tuned using the BFGS algorithm \cite{broyden1970convergence,mogensen2018optim}, improving predictive accuracy while retaining interpretability.

The SyTF framework controls model complexity through an adaptive parsimony mechanism. In the architecture, the expression complexity is measured by the number of nodes in the symbolic tree $\mathcal{E}$, denoted $C(\mathcal{E})$. To control overly complex mathematical functions, SyTF architecture incorporates an adaptive penalty that depends on how frequently and how recently expressions of each complexity level have appeared, a measure referred to as ‘frecency’. Thus, the modified loss function, following \cite{cranmer2023interpretable}, is defined as
$$
\ell(\mathcal{E})=\ell_{\text {pred }}(\mathcal{E}) \exp (\text {frecency }[C(\mathcal{E})]),
$$
where $\ell_{\text {pred }}(\mathcal{E})$ denotes the predictive loss and $\mathrm{frecency}\;[c]$ records how often and how recently expressions of complexity $c$ have appeared, computed with a moving time window and a tunable normalization constant. This adaptive mechanism dynamically regulates search pressure across complexity levels. When expressions of a particular complexity become overrepresented, their effective penalty increases, encouraging exploration of alternative structural forms. Conversely, underrepresented complexity levels incur a smaller penalty, making them more likely to be selected. This balanced exploration reduces the risk of early convergence to either overly simple or excessively complex formulas, thereby improving the likelihood of identifying forecasting models that achieve a practical balance between accuracy and interpretability. Additionally, SyTF adapts the \emph{Pareto front} strategy for model selection \cite{cranmer2023interpretable}. Instead of returning a single expression, it maintains a set of non-dominated solutions that balance predictive performance (loss) and model simplicity (expression complexity). An expression lies on the Pareto front if no other candidate is both more accurate and simpler. This multi-objective approach allows the selection of forecasting models that achieve an appropriate trade-off between interpretability and performance. 

In this study, we develop two variants of the SyTF framework: the base SyTF architecture employs core operators such as addition, subtraction, multiplication, and trigonometric functions ($\sin$ and $\cos$), whereas the SyTF-Div-Exp variant enlarges the operator set to include division and exponential functions. The restricted search space of the base model encourages simple expressions that align with the periodic and multiplicative patterns commonly observed in chaotic dynamics. In contrast, the division operator enables the extended variant to capture rational relationships prevalent in many physical and scientific systems, while the exponential function facilitates modeling of growth and decay processes. By comparing these two distinct configurations, we assess how a richer set of unary and binary operations influences both predictive performance and the interpretability of the resulting forecasting equations. Overall, by explicitly balancing forecasting performance and expression complexity, the SyTF framework produces human-readable mathematical expressions that provide transparent insights into the underlying temporal dynamics of the time series.

\begin{remark}
The proposed symbolic forecasting frameworks offer different trade-offs between model complexity, interpretability, and computational efficiency. In practice, the choice of model can be guided by the following considerations:
\begin{itemize}
    \item \textbf{SyNF and its variants.} The SyNF architecture is more suitable for datasets exhibiting complex nonlinear dynamics. These models tend to produce richer symbolic equations involving polynomial interactions and trigonometric components, allowing them to capture intricate temporal relationships. However, this increased expressive power results in higher computational complexity and longer training time. In practice, SyNF is often more effective for real-world datasets. Several variants further enhance the modeling capability: SyNF-Reg incorporates $\ell_1$ regularization to encourage sparsity and improve interpretability; SyNF-Div extends the operator library with division operators to represent rational functional relationships commonly observed in oscillatory or physical processes; and SyNF-Div-Reg combines division operations with $\ell_1$ regularization to balance expressive nonlinear modeling with sparse and interpretable symbolic equations.
    \item \textbf{SyTF and its variants.} The SyTF framework is preferred when the underlying time series dynamics are relatively simple or when computational efficiency is important. Due to the evolve-simplify-optimize learning procedure, SyTF typically produces compact symbolic expressions that resemble autoregressive structures. This makes it particularly suitable for simulated datasets or scenarios where simpler symbolic relationships are sufficient. The extended variant SyTF-Div-Exp further enriches the operator set by incorporating division and exponential functions, enabling the model to capture more complex nonlinear patterns while maintaining the compact symbolic structure of the SyTF framework.
    \end{itemize}
\end{remark}

\medskip
\noindent

\section{Experimental Setup and Results}\label{sec:experiments}

In this section, we evaluate the performance of the proposed symbolic forecasting approaches in predicting the future dynamics of both chaotic systems and real-world datasets. To establish a comprehensive comparison, the symbolic forecasters are benchmarked against a suite of state-of-the-art machine learning forecasting techniques. 

\subsection{Baseline Models} \label{Sec_Baseline_Models}
To ensure a systematic evaluation, we benchmark the symbolic forecasting methods against several forecasting frameworks spanning tree-based and deep learning techniques. This diverse selection of baseline models allows us to assess the robustness and generalization ability of symbolic methods across different modeling paradigms. In particular, we include several ensemble approaches such as the bagging-based Random Forest model \cite{breiman2001random}, the boosting-based Extreme Gradient Boosting (XGBoost) \cite{chen2016xgboost}, and the Light Gradient Boosting Machine (LightGBM) \cite{ke2017lightgbm}. These models are known for their ability to capture complex nonlinear relationships through aggregation and boosting strategies. From the deep learning paradigm, we consider several architectures that have demonstrated superior performance in sequence modeling tasks. We employ the Normalization-based Linear (NLinear) method \cite{zeng2023transformers}, which extends traditional single-hidden-layer neural networks by introducing normalization to stabilize learning dynamics and improve generalization.  We also use the Neural Basis Expansion Analysis for Time Series (NBetas) \cite{oreshkin2019n} and the Neural Hierarchical Interpolation for Time Series (N-HiTS) \cite{Challu2023} frameworks, which employ basis expansion and hierarchical decomposition to learn interpretable temporal patterns. In addition, we incorporate the recurrent architecture-based Long Short-Term Memory (LSTM) network \cite{hochreiter1997long}, which is well-suited for handling long-term temporal dependencies in time series data. Finally, to include attention-driven sequence modeling techniques, we evaluate the encoder-decoder frameworks, namely the Transformer model with a multi-head self-attention mechanism and the Time-series Dense Encoder (TiDE) \cite{das2023longterm}, which employs dense representations to enhance temporal encoding efficiency. Together, these models serve as robust baselines to comprehensively evaluate the efficacy of symbolic forecasting approaches across diverse datasets for one-step-ahead (rolling window) forecasting.

\subsection{Performance Indicators} \label{Sec_Performance_Metrics}
In our analysis, we evaluate the performance of symbolic forecasting approaches and baseline architectures using four scale-dependent and independent performance metrics, namely Symmetric Mean Absolute Percentage Error (SMAPE), Root Mean Squared Error (RMSE), Mean Absolute Error (MAE), and Mean Absolute Ranged Relative Error (MARRE). The mathematical formulation for these evaluation metrics is defined as follows:
$$
\text{SMAPE} = \frac{1}{h}\sum_{i=1}^{h} \frac{2\left|\hat{y}_{t+i} - y_{t+i}\right|}{\left(\left|\hat{y}_{t+i}\right| + \left|y_{t+i}\right|\right)} \times 100\%, \; \text{RMSE} = \sqrt{\frac{1}{h}\sum_{i=1}^{h} \left(\hat{y}_{t+i} - y_{t+i}\right)^2}, 
$$
$$
\text{MAE} = \frac{1}{h}\sum_{i=1}^{h} \left|\hat{y}_{t+i} - y_{t+i}\right|, \text{ and } \text{MARRE} = \frac{1}{h}\sum_{i=1}^{h} \left|\frac{y_{t+i} - \hat{y}_{t+i}}{\underset{i}{\max} y_{t+i} - \underset{i}{\min} y_{t+i}}\right|,
$$
where $y_{t+i} \in \mathbb{R}$ denotes the ground truth observation recorded at time $t+i$, $\hat{y}_{t+i} \in \mathbb{R}$ is the corresponding point forecast, and $h$ is the forecast horizon. By definition, smaller values of these metrics indicate better performance of the forecasting model \cite{hyndman2018forecasting, panja2023epicasting}.

\subsection{Forecast Evaluation for Chaotic Attractors} \label{Sec_Synthetic_Results}
We evaluate the performance of the symbolic forecasting approaches in predicting the future trajectories of 132 synthetic chaotic attractors. Our evaluation uses a rolling-window mechanism, where the model generates a one-step-ahead forecast for the predetermined test period. In this evaluation, four variants of symbolic forecasting techniques, including SyNF, its regularized variant SyNF-Reg, SyTF, and the extended SyTF-Div-Exp model that includes division and exponential operators to capture more complex nonlinear interactions, are considered. To ensure a comprehensive evaluation, all symbolic and baseline models are trained with different numbers of lagged observations (5, 10, and 25 lags), which serve as input features providing temporal context. The smaller number of lags enhances model interpretability and computational effectiveness, while a larger number of lags offers richer historical information and yields complex representations. 


\begin{figure}
    \centering
    \includegraphics[width=\linewidth]{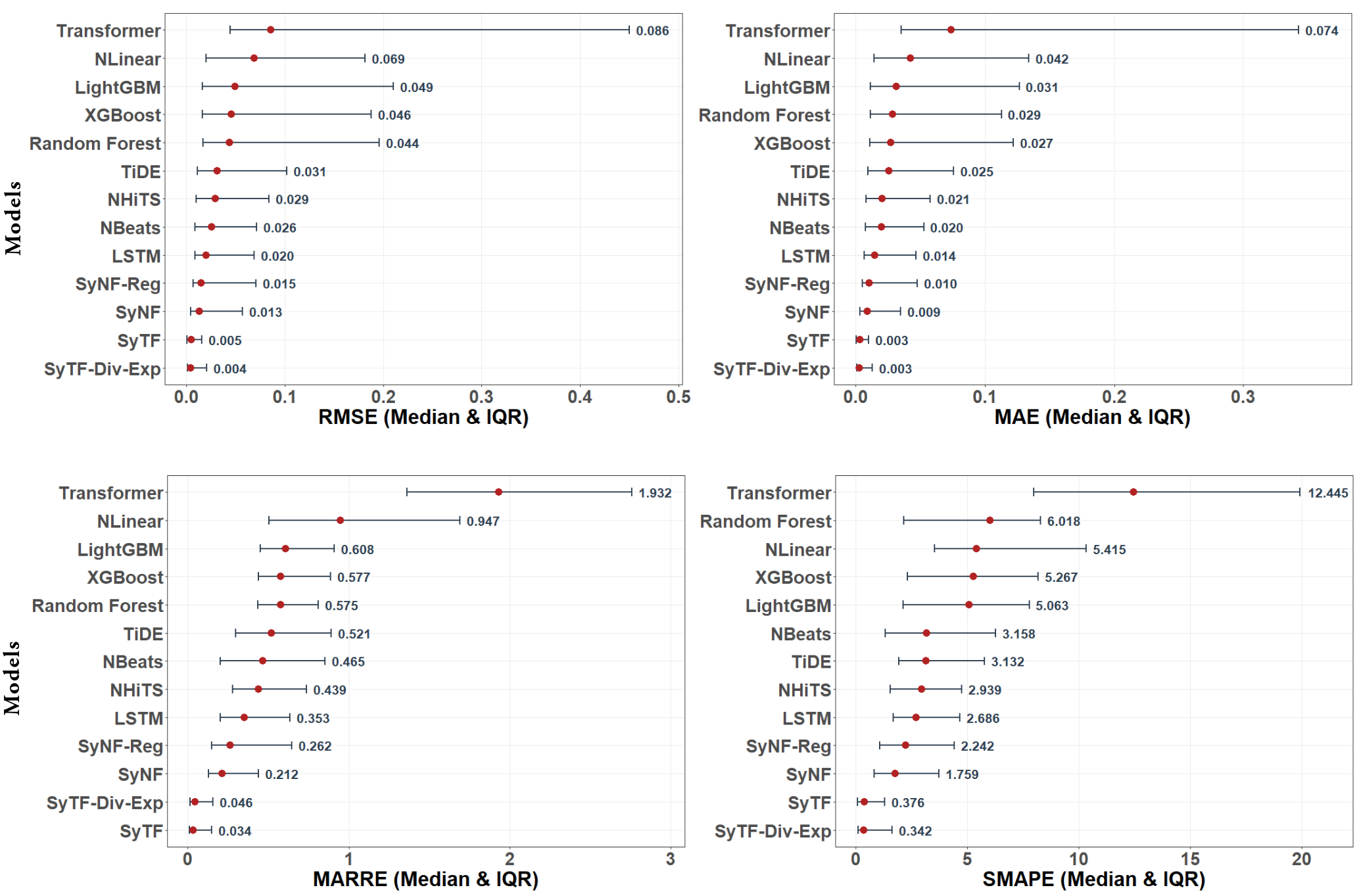}
    \caption{Comparative performance of symbolic forecasting approaches and state-of-the-art frameworks for one-step ahead (rolling window) forecasting of chaotic datasets using \textbf{five} historical observations for each model. In the plot, panels show error distributions across four metrics: RMSE (top-left), MAE (top-right), MARRE (bottom-left), and SMAPE (bottom-right). Red dots denote median performance; error bars represent the interquartile range (IQR). For all panels, models are ranked by ascending median error (lower values indicate better performance).}
    \label{EQL_PySR_Lag_5}
\end{figure}

The empirical results, illustrated in Figure \ref{EQL_PySR_Lag_5}, summarize the forecasting performance of all models trained with five lagged observations. The errorbar plots display the median and interquartile range of four key performance metrics across the chaotic attractor datasets. The results highlight the superiority of symbolic forecasting architectures, particularly with the genetic programming-based SyTF-Div-Exp and SyTF models consistently achieving the minimum median errors and dispersion, reflecting both high accuracy and stability across attractors. These expression tree-based forecasters excel by discovering explicit analytical expressions that approximate the governing dynamics of chaotic systems, yielding interpretable and generalizable predictive relations. The SyNF and its regularized variant, which integrate symbolic operations with a neural training mechanism, achieve competitive yet less stable performance compared to SyTF. While the gradient-based optimization in SyNF models enhances flexibility and smoother learning dynamics, the stochastic nature of neural training introduces variability that reduces their generalization capability across chaotic regimes. Despite their higher variability, both SyNF variants consistently outperform state-of-the-art deep learning architectures in forecasting performance. Among the baseline models, deep neural forecasters such as NBeats, NHiTS, TiDE, and LSTM showcase moderate median performance but fail to generalize effectively to the inherent instability of chaotic systems. On the other hand, the tree-based ensemble methods like XGBoost, LightGBM, and Random Forest exhibit larger variance, while the attention mechanism-based Transformer framework records both higher median errors and wider error dispersion primarily due to the low sample size of the dataset. 

\begin{figure}
    \centering
    \includegraphics[width=\linewidth]{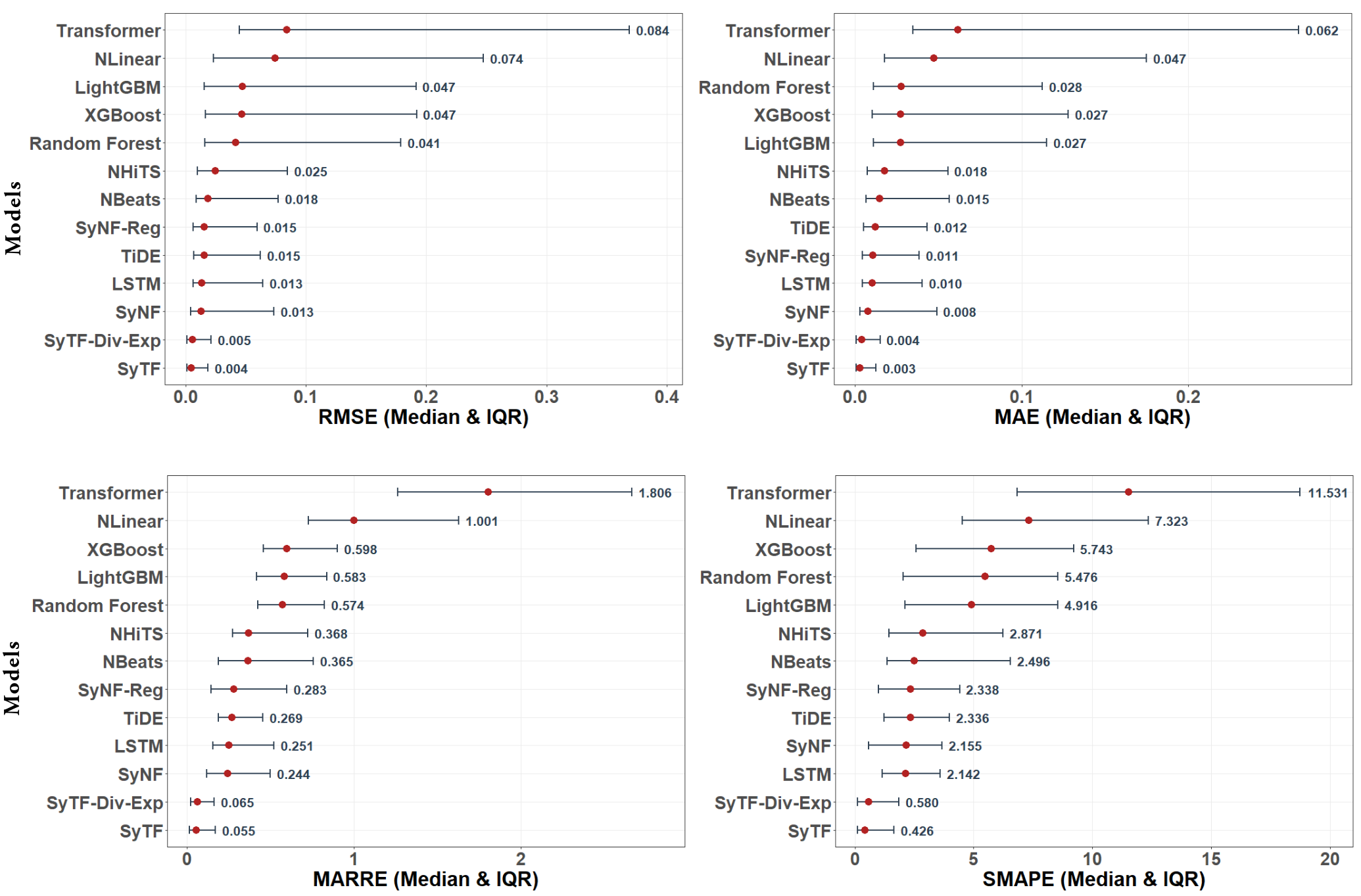}
    \caption{Comparative performance of symbolic forecasting approaches and state-of-the-art frameworks for one-step ahead forecasting of chaotic datasets using \textbf{ten} historical observations for each model. In the plot, panels show error distributions across four metrics: RMSE (top-left), MAE (top-right), MARRE (bottom-left), and SMAPE (bottom-right). Red dots denote median performance; error bars represent the interquartile range (IQR). For all panels, models are ranked by ascending median error (lower values indicate better performance).}
    \label{EQL_PySR_Lag_10}
\end{figure}

\begin{figure}
    \centering
    \includegraphics[width=\linewidth]{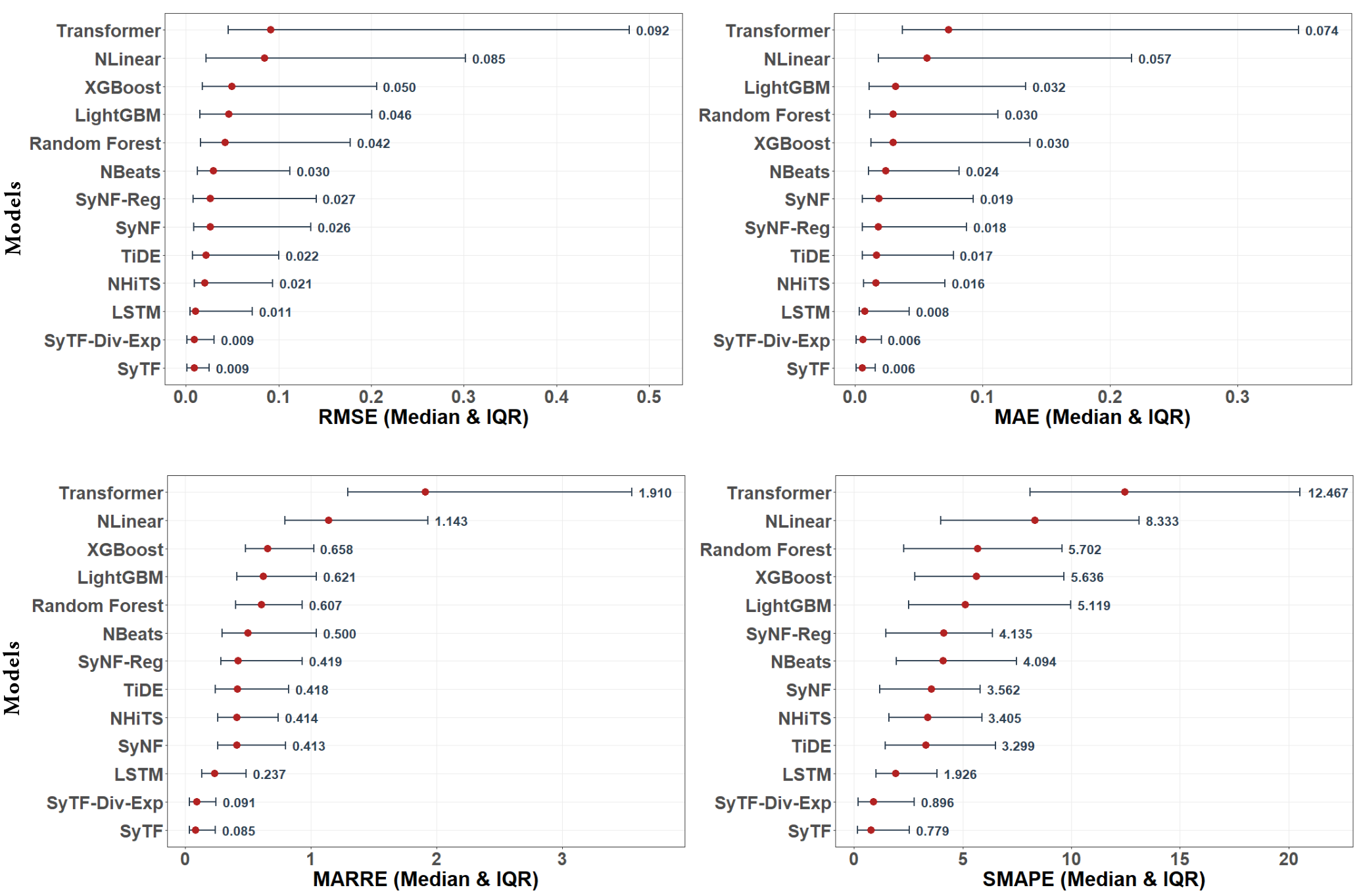}
    \caption{Comparative performance of symbolic forecasting approaches and state-of-the-art frameworks for one-step ahead forecasting of chaotic datasets using \textbf{twenty-five} historical observations for each model. In the plot, panels show error distributions across four metrics: RMSE (top-left), MAE (top-right), MARRE (bottom-left), and SMAPE (bottom-right). Red dots denote median performance; error bars represent the interquartile range (IQR). For all panels, models are ranked by ascending median error (lower values indicate better performance).}
    \label{EQL_PySR_Lag_25}
\end{figure}

\paragraph{Ablation Study.} Additionally, we conduct an ablation study to determine the influence of the number of lagged observations on the performance of the symbolic forecasting approaches. The emipircal evaluations conducted with 10 and 25 lagged observations, presented in Figures  \ref{EQL_PySR_Lag_10} and \ref{EQL_PySR_Lag_25}, also depict similar patterns. While increasing the lag window enhances the temporal representation and allows the models to capture long-term dependencies, the relative ranking among the models remains consistent. The SyTF architecture and its extended variant exhibit the lowest median forecasting errors and highest stability across all chaotic attractors. This demonstrates the robustness of the SyTF frameworks to changes in input dimensionality and the ability to optimize model performance and complexity. The SyNF and SyNF-Reg models also maintain competitive performance with the SyTF frameworks. On the other hand, the deep learning forecasters depict marginal improvements in performance due to additional lags; however, they continue to suffer from increased error dispersion and reduced generalization. The tree-based models and other baseline architectures demonstrate a similar pattern of limited adaptability as the input horizon expands.

\begin{figure}
    \centering
    \includegraphics[width=1.0\linewidth]{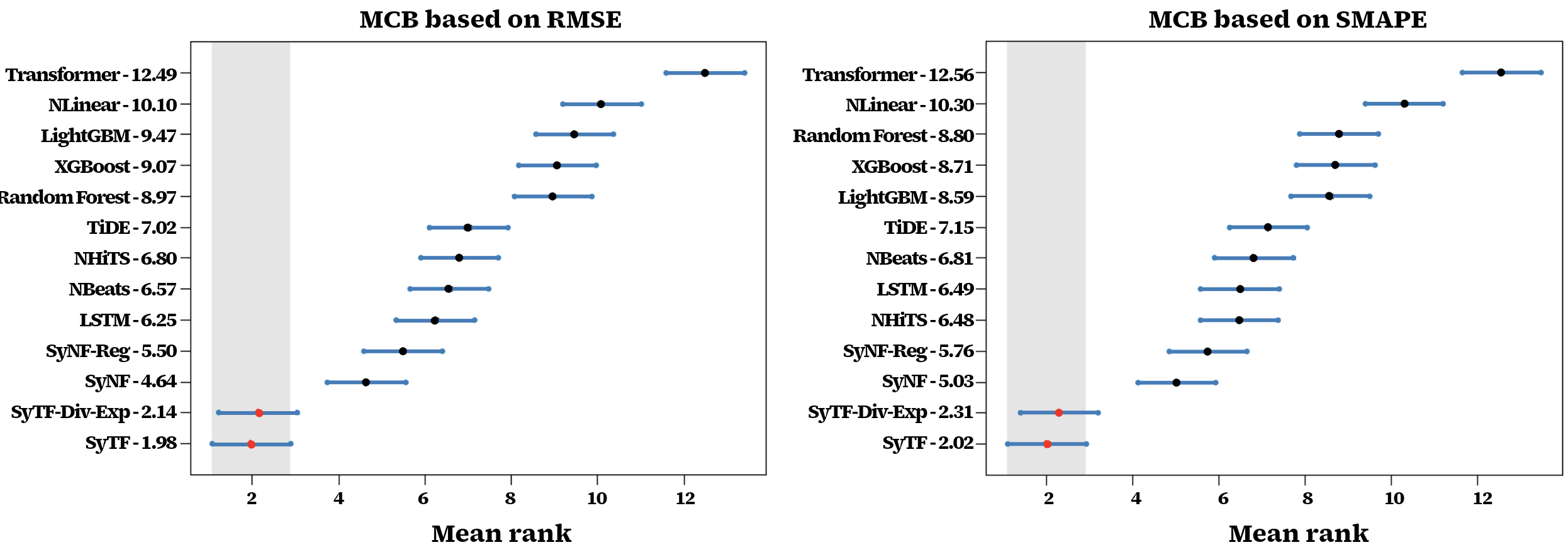}
    \caption{Visualization of the MCB test results for synthetic chaotic attractor forecasting tasks based on RMSE (left) and SMAPE (right) metrics. The vertical axis represents the average ranks of the forecasting methods, while the horizontal axis lists the corresponding techniques. For example, {SyTF – 1.98} indicates that the SyTF framework obtained an average rank of 1.98 under the RMSE metric, with similar interpretations applicable to the remaining models.}
    \label{Fig_MCB_Plot}
\end{figure}

\paragraph{Statistical Significance Test.} Furthermore, to validate the robustness of the performance improvements of the SyTF and SyTF-Div-Exp approach in forecasting the dynamics of the chaotic attractors, we conduct the Multiple Comparison with the Best (MCB) test \cite{nemenyi1963distribution}. This non-parametric test procedure ranks the competitive models based on their performance across different forecasting tasks and computes the average rank with their critical distance. The model with the least average rank is considered the `best' performing technique, and the corresponding critical distance serves as the reference value for the test. Figure \ref{Fig_MCB_Plot} presents the MCB test results, in terms of RMSE and SMAPE metrics, for the symbolic and baseline forecasting models with five lagged inputs. As observed from the results, the SyTF model achieves the minimum ranks in both the performance metrics and hence serves as the `best' forecasting model, followed by the SyTF-Div-Exp framework. Since the critical distance of all other models lies beyond the reference value (shaded region) of the test, their performance is significantly inferior in comparison to the SyTF architecture. Overall, the empirical evaluation on the chaotic attractor datasets underscores the superior performance of symbolic forecasting frameworks, where SyTF and SyTF-Div-Exp emerge as the most accurate and stable architectures, while the SyNF variants offer a robust balance between symbolic representation and neural adaptability, outperforming all baselines. Beyond their predictive performance, these symbolic models possess the critical benefit of interpretability, providing analytical relationships that explain the underlying chaotic dynamics, unlike conventional deep learning approaches, which function as black-box architectures.

\subsection{Empirical Evaluation for Real-world Dataset} \label{Sec_Realworld_Results}
We assess the forecasting performance of the proposed symbolic forecasting frameworks on two complex real-world time series datasets, San Juan dengue and El Niño SST, as described in Section \ref{Sec_Real_World_Time_Series}. These datasets represent diverse real-world phenomena, encompassing epidemiological and climatic domains, each characterized by nonlinear, nonstationary, and chaotic dynamics. In addition to the competitive benchmark models discussed earlier, we evaluate two enhanced variants of the proposed SyNF framework: (i) SyNF-Div, which incorporates division-based symbolic operators for richer functional expressiveness, and (ii) SyNF-Div with $\ell_1$ regularization (SyNF-Div-Reg), which introduces sparsity constraints to prevent overfitting and enhance model interpretability. For consistency and fair comparison, we employ a comprehensive experimental setup where four historical observations of the target variable are used as inputs to predict one-step-ahead values over the test horizon across all forecasting techniques. This choice is motivated by the short-memory behavior of these time series, as evident from the preliminary analysis (Figure \ref{table_acf_pacf}), which indicates that only a few recent observations are sufficient to capture the relevant temporal dependencies. 

\begin{table}[!ht]
\caption{Comparative evaluation of forecasting performance for machine learning models and symbolic forecasting frameworks on real-world datasets across multiple performance indicators. In the table, the \textbf{\underline{best}} and \textit{second-best} performing models are highlighted.}
    \centering
    \footnotesize
    \begin{tabular}{c|cccc|cccc}
    \hline
        {Dataset} & \multicolumn{4}{c|}{Sanjuan Dengue} & \multicolumn{4}{c}{El Niño SST} 
        \\ \hline
        Models & {SMAPE} & {RMSE} & {MAE} & {MARRE} & {SMAPE} & {RMSE} & {MAE} & {MARRE}  
        \\ \hline
        NLinear & 21.70 & 21.37 & 16.97 & 7.44 & 1.03 & 0.34 & 0.27 & 8.29 
        \\ 
        RandomForest & 25.21 & 25.46 & 20.08 & 8.81 & 0.87 & 0.30 & 0.22 & 6.94 
        \\ 
        XGBoost & 27.58 & 28.09 & 22.16 & 9.72 & 0.92 & 0.31 & 0.24 & 7.29 
        \\ 
        LightGBM & 24.27 & 32.62 & 21.99 & 9.64 & \textit{0.82} & 0.28 & \textit{0.21} & \textit{6.55} 
        \\ 
        LSTM & 50.21 & 72.43 & 49.98 & 21.92 & 5.23 & 1.62 & 1.40 & 42.41 
        \\ 
        NBeats & 21.77 & 21.94 & 16.84 & 7.38 & 0.83 & 0.29 & \textit{0.21} & 6.61 
        \\ 
        NHiTS & 21.10 & 21.59 & 16.29 & 7.14 & 0.86 & 0.30 & 0.22 & 6.87 
        \\ 
        Transformer & 81.54 & 94.41 & 70.11 & 30.75 & 4.36 & 1.38 & 1.15 & 35.12 
        \\
        TiDE & 21.94 & 22.18 & 17.14 & 7.51 & 1.19 & 0.37 & 0.32 & 9.59 
        \\ \hline
        SyNF & 21.39 & \textit{20.80} & 16.19 & 7.10 & 0.83 & \textit{0.27} & 0.22 & 6.66 
        \\ 
        SyNF-Reg & \textbf{\underline{20.95}} & \textbf{\underline{20.69}} & \textbf{\underline{15.76}} & \textbf{\underline{6.91}} & 1.41 & 0.44 & 0.37 & {11.37} 
        \\ 
        SyNF-Div & \textit{21.09} & 20.82 & \textit{16.04} & \textit{7.03} & 0.84 & \textit{0.27} & 0.22 & 6.70 
        \\ 
        SyNF-Div-Reg & 21.36 & 20.96 & 16.60 & 7.28 & \textbf{\underline{0.79}} & \textbf{\underline{0.26}} & \textbf{\underline{0.20}} & \textbf{\underline{6.30}} 
        \\ 
        SyTF & 23.30 & 23.32 & 17.76 & 7.79 & 0.86 & \textit{0.27} & 0.22 & 6.88 
        \\
        SyTF-Div-Exp & 22.73 & 24.04 & 18.37 & 8.06 & 0.85 & \textit{0.27} & 0.22 & 6.79 
        \\ \hline
    \end{tabular}
    \label{Table_Real_World_Performance_Comparison}
\end{table}

\begin{figure}
    \centering
    \includegraphics[width=1.0\linewidth]{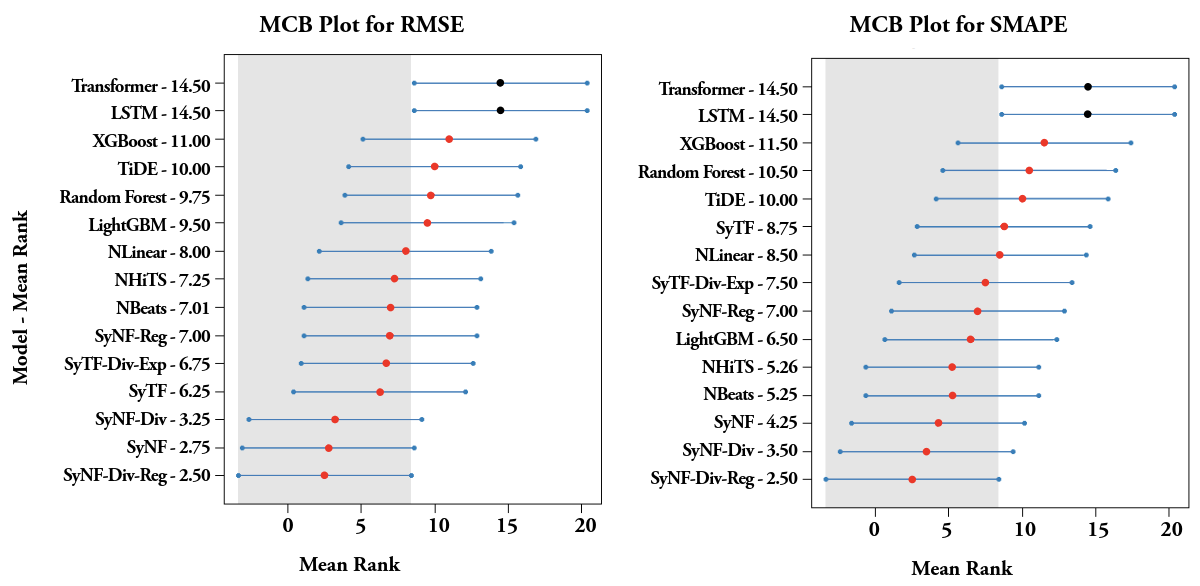}
    \caption{Visualization of the MCB test results for real-world forecasting tasks based on RMSE (left) and SMAPE (right) metrics. The vertical axis represents the average ranks of the forecasting methods, while the horizontal axis lists the corresponding techniques. For instance, {SyNF-Div-Reg – 2.50} denotes that the SyNF-Div-Reg framework achieved an average rank of 2.50 according to the RMSE metric (lower metric indicates better performance), and similar interpretations apply to the other methods.}
    \label{Real_World_MCB_Figure}
\end{figure}

Table \ref{Table_Real_World_Performance_Comparison} presents a comprehensive performance evaluation for one-step ahead forecasting of real-world datasets for all competing models during the test period. For the San Juan dengue dataset, conventional machine learning approaches such as Random Forest, XGBoost, and LightGBM yield moderate performance but struggle to capture sudden epidemic outbreaks, resulting in higher RMSE and MAE values compared to neural forecasting models. Deep learning architectures such as NBeats, N-HiTS, and TiDE provide competitive performance, while recurrent models such as LSTM and attention-based Transformers exhibit significantly higher errors due to their inability to model irregular epidemic patterns with limited data. In contrast, the SyNF-Reg architecture achieves the lowest errors across all evaluation metrics, indicating that the sparse symbolic connections effectively capture the inherent nonlinear patterns of dengue incidence cases. Notably, SyNF and SyNF-Div also outperform most deep learning baselines, suggesting that symbolic representations provide a more data-efficient mechanism for modeling epidemiological time series than conventional neural architectures. For the El Niño SST dataset, characterized by complex oscillatory behavior and long-range temporal dependencies, the forecasting performance differs across model paradigms. While several neural forecasting architectures, such as NBeats, N-HiTS, and LightGBM, achieve relatively low errors, the performance of recurrent and attention-based models consistently diminishes. Notably, the SyNF-Div-Reg model outperforms all baseline frameworks, achieving the lowest SMAPE, RMSE, MAE, and MARRE metrics for this dataset. The inclusion of division operations combined with the sparse connections enables the model to reconstruct complex SST fluctuations by representing rational functional relationships, which are particularly effective for describing oscillatory physical processes. Other symbolic forecasting approaches also remain highly competitive, consistently performing similar or better than the state-of-the-art forecasting model. Additionally, we present the explicit analytical expressions learned by the symbolic forecasting frameworks in Table \ref{tab:symbolic_eq_real}. These symbolic equations describe the temporal dynamics of the real-world datasets as modeled by the SyNF and SyTF architectures. Compared to SyNF, the equations obtained from the SyTF model are considerably simpler and largely resemble compact autoregressive relationships involving only a few lagged observations. This simplicity can be attributed to the evolve-simplify-optimize loop in SyTF, which balances exploration and exploitation while capturing most temporal dependencies in the latent representation, allowing the symbolic layer to remain concise. In contrast, the SyNF model produces richer nonlinear expressions involving polynomial interactions and trigonometric terms. For the El Niño SST dataset, these periodic components reflect the oscillatory nature of sea surface temperature dynamics, while for the San Juan Dengue dataset, they help in capturing nonlinear seasonal patterns in epidemic incidence. Overall, these symbolic equations provide interpretable insights into the dominant temporal dependencies governing the evolution of the underlying processes. Furthermore, to validate the statistical significance of the observed performance improvements, we conduct the distribution-free MCB test. The results of this test, illustrated in Figure \ref{Real_World_MCB_Figure}, indicate that the SyNF-Div-Reg framework consistently outperforms competitive deep learning and ensemble models across both real-world forecasting tasks, followed by the SyNF and SyNF-Div models. 

\begin{table}[]
    \centering
    \caption{Symbolic equations learned by SyTF and SyNF for real-world datasets.}
    \scriptsize
    \begin{tabular}{c|c|c}
    \hline
    \multirow{2}{*}{Dataset}& \multicolumn{2}{c}{Equation from} \\ \cline{2-3}
    & SyTF $\left(\hat{y}_{t, \text{SyTF}} = \right)$ & SyNF $\left(\hat{y}_{t} = \right)$ \\ \hline
    Sanjuan
    & $y_{t-4} 0.087 \sin(y_{t-2} +$ 
    & $-1.1e^{-4} y_{t-1}^2 + 6.7e^{-5}  y_{t-1} y_{t-2} + 0.288 y_{t-1} + 3.7e^{-5} y_{t-1}^2 + 0.684 y_{t-2}  + 0.938 -$  \\ 
    Dengue & $\sin(0.280 y_{t-3} - 0.696)) +y_{t-4} $ 
    & $0.203\sin(-3.6e^{-4} y_{t-1} + 1.337 y_{t-2} + 1.439) + 0.601\cos(1.568 y_{t-1} - 0.288 y_{t-2} + 0.063)$ \\ \hline
    El Niño 
    & \multirow{2}{*}{$0.985  y_{t-1} + 0.417$} 
    & $-8.2e^{-3} y_{t-1}^2 + 0.017 y_{t-1} y_{t-2} + 0.553 y_{t-1} + 0.019 y_{t-2}^2 - 0.324 y_{t-2} + 0.031 -$ \\
    SST & &  $ 0.137\sin(1.438 y_{t-1} + 0.078 y_{t-2} - 0.107) + 1.792\cos(0.536 y_{t-1} + 3.2e^{-5} y_{t-2} - 0.463) $ \\ \hline    
    \end{tabular}
    \label{tab:symbolic_eq_real}
\end{table}


Overall, the empirical findings for the real-world datasets underscore the superior generalization ability of the symbolic neural architectures (SyNF family) compared to the expression-tree-based symbolic tree forecasters. The performance improvements observed in SyNF arise from its hybrid design, which effectively integrates symbolic reasoning with neural parameter optimization. Unlike conventional neural networks, SyNF maintains a fixed computational structure where only the connection weights between the input and symbolic nodes are optimized through gradient-based learning. This design ensures both computational efficiency and symbolic interpretability, enabling the model to capture complex nonlinear dependencies of real-world time series observations while remaining structurally stable. In contrast, the SyTF framework employs an evolutionary symbolic search mechanism that iteratively constructs expression trees, allowing both the symbolic structure and associated parameters to evolve over time. Although this approach provides flexibility and can achieve strong performance in low-dimensional and noise-free settings, its effectiveness diminishes for high-dimensional, noisy, and chaotic datasets. The rapidly expanding symbolic search space under these conditions leads to computational inefficiency and suboptimal convergence, ultimately degrading forecasting performance for real-world datasets.

\subsection{Uncertainty Quantification}\label{Sec_Uncertainty_Quantification}
\begin{figure}
    \centering
    \includegraphics[width=1.0\linewidth]{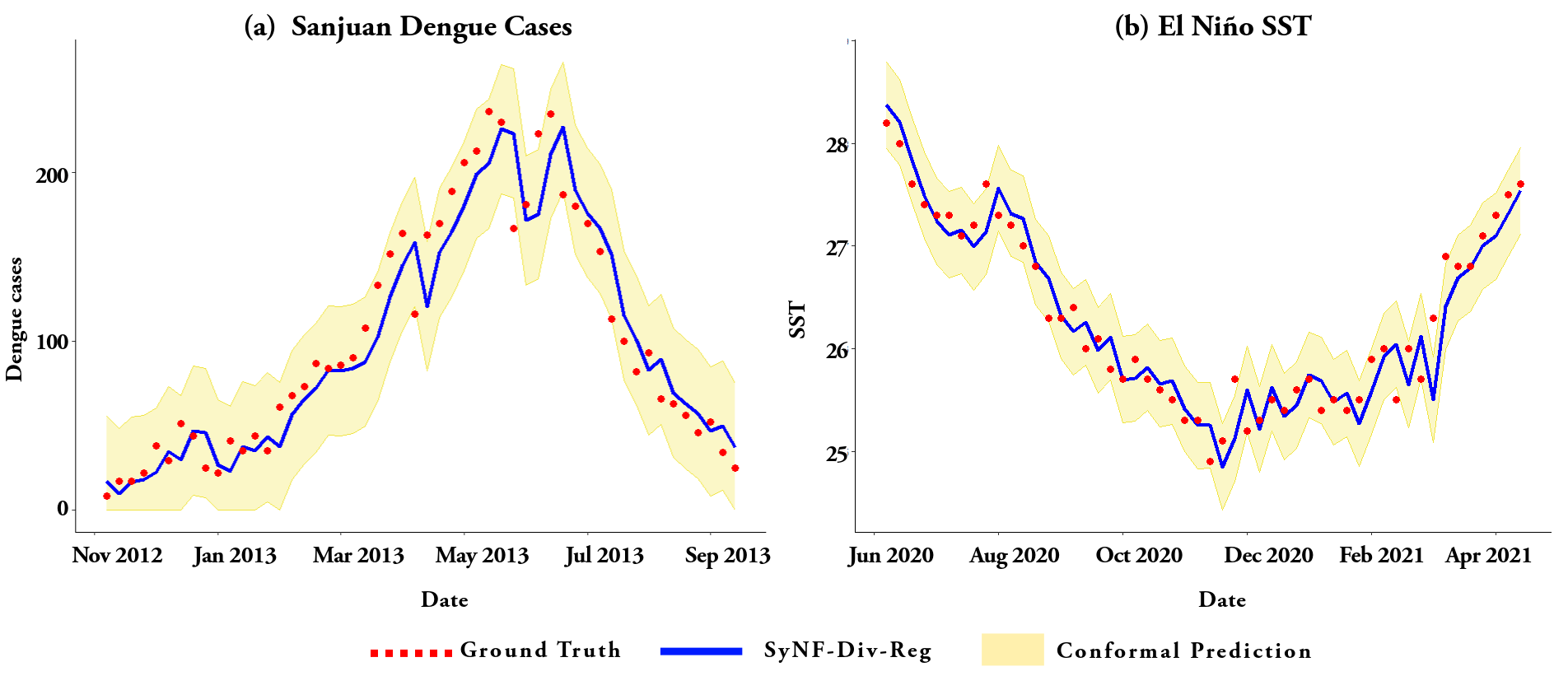}
    \caption{The plot visualizes the ground truth (red points) observations of (a) Sanjuan Dengue and (b) El Niño SST dataset during the test period, along with the point forecasts (blue line) and the conformal prediction interval of the SYNF-Div-Reg model (yellow shaded).}
    \label{Fig_CP_Plot}
\end{figure}

Alongside point forecasts, we also applied a conformal prediction approach to quantify the uncertainty in the forecasts produced by the symbolic forecasting models. This non-parametric method offers a systematic way to transform deterministic predictions from the symbolic forecasters ($\mathcal{SF}$) into reliable prediction intervals \cite{vovk2005conformal}. Given the input sequence ${y_t}$, the approach constructs an uncertainty model ($\mathcal{UM}$) based on the lagged inputs $\left\{\underline{y}_{t-1}\right\}$ to capture the predictive uncertainty. Using this setup, a conformal score ($\mathcal{CS}_t$) is then derived as
$$
    \mathcal{CS}_t = \frac{\left|y_t - \mathcal{SF}\left(\underline{y}_{t-1}\right)\right|}{\mathcal{UM}\left(\underline{y}_{t-1}\right)}.
$$
The model-agnostic conformal prediction approach leverages the inherent temporal patterns of the input series and computes the conformal quantile $\left(\mathcal{CQ}_t\right)$ using a weighted conformal method with an $\alpha$-sized window $\Gamma_t = \mathbf{1}\left(\tilde{t} \geq t - \alpha\right), \forall \; \tilde{t} < t$ as 
\begin{equation*}
    \mathcal{CQ}_t = \inf\left\{\Delta: \frac{1}{\min\left(\alpha, \tilde{t} -1\right) + 1} \sum_{\tilde{t} = 1}^{t-1} \mathcal{CS}_{\tilde{t}} \Gamma_{\tilde{t}} \geq 1- \delta \right\}.
\end{equation*}
Hence, the $100*\left(1- \delta\right)\%$ conformal prediction interval based on these weighted quantiles is given by:
\begin{equation*}
    \mathcal{SF}\left(\underline{y}_{t-1}\right) \pm \mathcal{CQ}_t \; \mathcal{UM}\left(\underline{y}_{t-1}\right).
\end{equation*}
In this analysis, we construct 90\% conformal prediction intervals for the real-world forecasting tasks generated by the best-performing SyNF-Div-Reg model, as identified in Figure \ref{Real_World_MCB_Figure}. To ensure the statistical validity of these intervals and prevent any potential data leakage, the conformal residuals are derived using a separate calibration (validation) set, distinct from the test period. Figure \ref{Fig_CP_Plot} illustrates both the point forecasts produced by the SyNF-Div-Reg model and the corresponding ground-truth observations, with shaded regions representing the conformal prediction intervals. As observed, the SyNF-Div-Reg framework effectively captures the intricate temporal patterns and nonlinear fluctuations inherent in both the San Juan dengue incidence and El Niño SST datasets. The predicted trajectories align closely with the observed values, and the majority of data points lie within the conformal intervals, underscoring the model’s predictive reliability. Notably, the intervals adapt dynamically to the underlying data regimes, expanding during highly volatile periods and contracting in stable phases, demonstrating the ability of SyNF-Div-Reg to represent regime-dependent uncertainty and temporal heterogeneity in real-world dynamics. Overall, the SyNF-Div-Reg model provides the best performance in both point forecasting and uncertainty estimation for real-world chaotic time series datasets. 

\section{Conclusion and Future Work}
The integration of symbolic discovery with contemporary forecasting pipelines offers a pathway toward models that are not only accurate but also analytically tractable and scientifically meaningful. This study provides the first large-scale benchmark for symbolic machine learning in chaotic time-series forecasting, positioning symbolic model discovery as a viable approach to interpretable forecasting of chaotic dynamics. Across 132 chaotic attractors, expression-tree methods, such as SyTF and its richer operator variant, achieve the most accurate and stable one-step-ahead forecasts, outperforming a wide range of deep learning and ensemble baselines. SyTF also yields compact equations that expose the inherent patterns of the underlying dynamics. 

On real-world data, the strongest performance shifts toward the neural-symbolic family: sparsity and rational operators improve generalization. In particular, regularized SyNF variants performed best for dengue incidence, and the division-regularized variant performed best for Niño 3.4 SST. Beyond point accuracy, the symbolic neural framework supports uncertainty quantification through conformal prediction intervals that adapt to volatility, offering a practical pathway toward reliable, interpretable forecasting in high-stakes applications.

Several extensions follow naturally. First, while we focus on rolling-window nowcasting, an important next step is to evaluate multi-step-ahead performance, including the stability of discovered equations under iterative forecasting. Second, incorporating multivariate inputs and exogenous drivers (especially for real-world climate and epidemic series) could improve both the accuracy and interpretability of the recovered relationships. Third, tighter integration between symbolic discovery and dynamical system constraints, such as stability or invariance priors, may further reduce the expression search space while improving robustness. Although this work represents an initial attempt to integrate symbolic machine learning techniques for forecasting real-world time series, future research should also explore their application in high-risk domains such as mobile health monitoring, and continuous time series from electronic health records including heart rate, ECG, and EEG signals, where the interpretability of forecasting models is critically important. We hope this benchmark stimulates further research at the intersection of forecasting, dynamical systems, and interpretable machine learning.

\section*{Data and Code Availability Statement}
Data and codes are available in our GitHub repository: \url{https://github.com/mad-stat/Symbolic_Forecasting}.

\bibliographystyle{plain}
\bibliography{references}

\end{document}